%% file: main.tex
\documentclass[10pt,twocolumn,letterpaper]{article}

\usepackage[pagenumbers]{cvpr} 

\usepackage{graphicx}
\usepackage{amsmath}
\usepackage{amssymb}
\usepackage{booktabs}
\usepackage{amsfonts}
\usepackage{bbding} 
\usepackage{color}
\usepackage{multirow}
\usepackage{multicol}
\usepackage[accsupp]{axessibility}
\usepackage{makecell}
\usepackage{colortbl} 
\usepackage{caption}
\definecolor{Gray}{gray}{0.92} 

\usepackage{graphics} 
\usepackage{epsfig} 
\usepackage{mathptmx} 
\usepackage{times} 
\usepackage{amsmath} 
\usepackage{amssymb}  
\usepackage{siunitx}  
\usepackage{newtxtext,newtxmath} 

\usepackage{times}  
\usepackage{helvet}  
\usepackage{courier}  
\usepackage[hyphens]{url}  
\usepackage{graphicx} 
\usepackage{caption} 
\usepackage{amsmath}
\usepackage{amssymb}
\usepackage{colortbl}
\usepackage{subcaption}
\usepackage{booktabs} 
\usepackage{multirow}
\usepackage[dvipsnames, svgnames, x11names]{xcolor}
\definecolor{lgray}{gray}{0.7}
\usepackage{algorithm}
\usepackage{algorithmic}
\usepackage{enumitem}
\usepackage{bibentry}

\usepackage{newfloat}
\usepackage{listings}

\definecolor{cvprblue}{rgb}{0.21,0.49,0.74}

\usepackage[pagebackref,breaklinks,colorlinks]{hyperref}

\usepackage[capitalize]{cleveref}
\crefname{section}{Sec.}{Secs.}
\Crefname{section}{Section}{Sections}
\Crefname{table}{Table}{Tables}
\crefname{table}{Tab.}{Tabs.}

\def\cvprPaperID{0001} 
\def\confName{CVPR}
\def\confYear{2024}

\DeclareUnicodeCharacter{0301}{\'{e}}

\begin{document}

\title{U-ViLAR: Uncertainty-Aware Visual Localization for Autonomous \\Driving via Differentiable Association and Registration}

\author{
Xiaofan Li\textsuperscript{*},
Zhihao Xu\textsuperscript{*},
Chenming Wu,
Zhao Yang,
Yumeng Zhang,
Jiang-Jiang Liu, \\
Haibao Yu,
Fan Duan,
Xiaoqing Ye,
Yuan Wang,
Shirui Li,
Xun Sun,
Ji Wan,
Jun Wang \\
\vspace{4mm}
\textsuperscript{} Baidu Inc.\\
}

\maketitle

\renewcommand{\thefootnote}{\fnsymbol{footnote}}
\footnotetext[1]{Equal contribution.}


\input{sec/0_abstract}

\input{sec/1_intro}

\input{sec/2_related}

\input{sec/3_method}

\input{sec/4_exp}
\input{sec/5_conclusion}

{\small
\bibliographystyle{ieee_fullname}
\bibliography{main}
}
\end{document}

%% file: sec/0_abstract.tex
\begin{abstract}
Accurate localization using visual information is a critical yet challenging task, especially in urban environments where nearby buildings and construction sites significantly degrade GNSS (Global Navigation Satellite System) signal quality. This issue underscores the importance of visual localization techniques in scenarios where GNSS signals are unreliable. This paper proposes U-ViLAR, a novel uncertainty-aware visual localization framework designed to address these challenges while enabling adaptive localization using high-definition (HD) maps or navigation maps.  
Specifically, our method first extracts features from the input visual data and maps them into Bird’s-Eye-View (BEV) space to enhance spatial consistency with the map input. Subsequently, we introduce:  
\textbf{a)} \textbf{Perceptual Uncertainty-guided Association}, which mitigates errors caused by perception uncertainty, and  
\textbf{b)} \textbf{Localization Uncertainty-guided Registration}, which reduces errors introduced by localization uncertainty.
By effectively balancing the coarse-grained large-scale localization capability of association with the fine-grained precise localization capability of registration, our approach achieves robust and accurate localization. 
Experimental results demonstrate that our method achieves state-of-the-art performance across multiple localization tasks.
Furthermore, our model has undergone rigorous testing on large-scale autonomous driving fleets and has demonstrated stable performance in various challenging urban scenarios.

\end{abstract}

%% file: sec/1_intro.tex
\begin{figure}[t]
\centering
\includegraphics[width=\linewidth]{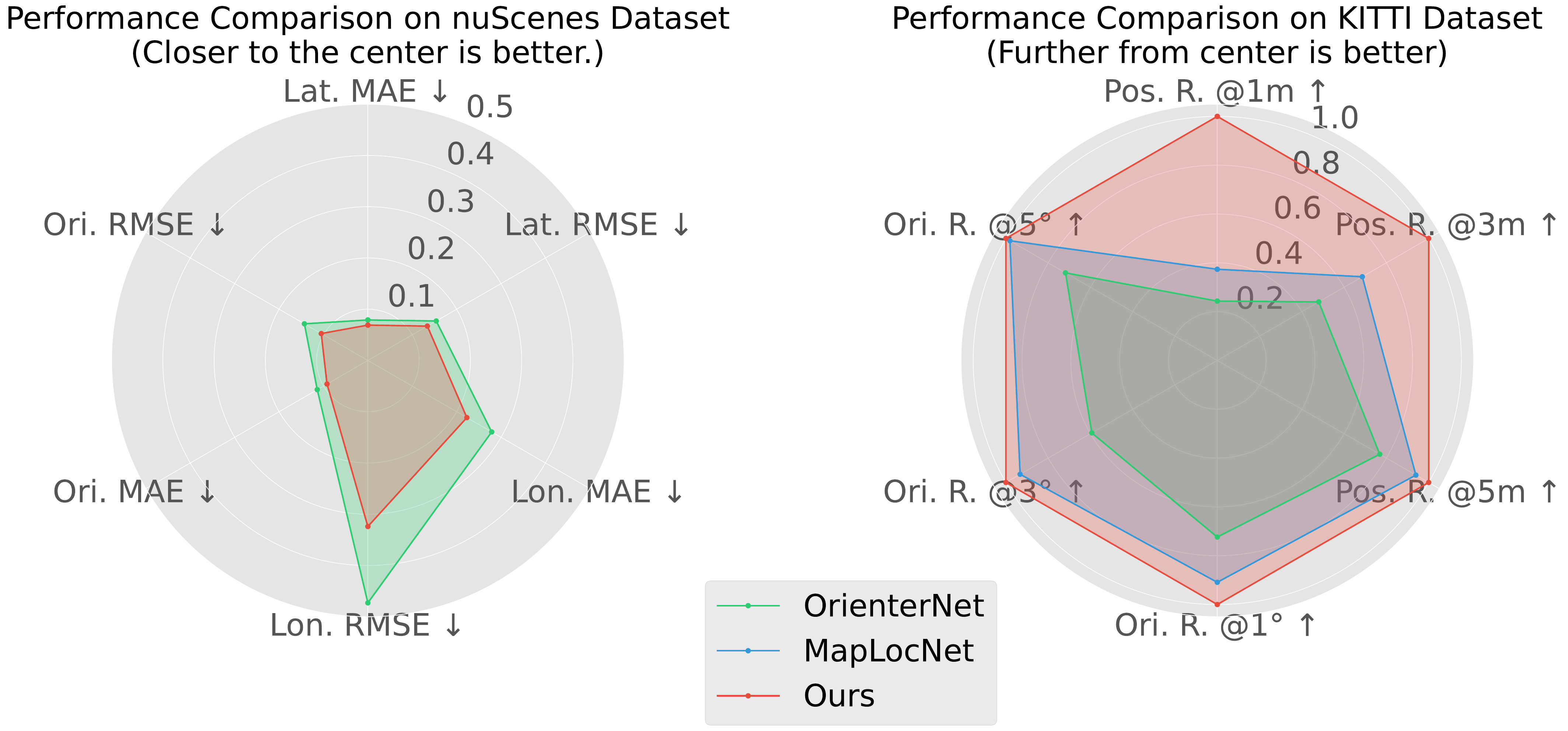}
\caption{
In \textbf{Fine-grained Localization (left)} and \textbf{Large-scale Relocalization (right)}, U-ViLAR outperforms all existing methods across all metrics. Lat.', Lon.', and Ori.' represent Lateral, Longitudinal, and Orientation, respectively, while R.' denotes Recall. }
\centering
\vspace{-6mm}
\label{fig:m2}
\end{figure}

\section{Introduction}
Autonomous vehicles rely significantly on Global Navigation Satellite Systems (GNSS) for outdoor localization. However, GNSS signals are susceptible to noise caused by obstructions from buildings, tunnels, and bridges, which complicates the accuracy of GNSS-based localization. In this context, visual localization using imprecise GNSS data is becoming increasingly critical.
In autonomous driving systems, localization typically demands centimeter-level accuracy~\cite{Brsan2018LearningTL}, whether using high-definition (HD) maps or lightweight navigation maps. Both types of maps are crucial for autonomous driving due to their precision or cost-effectiveness. In this regard, an end-to-end visual localization system that can adapt to both map types and coarse GNSS signals is essential.

Classical methods for visual localization rely on establishing 2D-3D correspondences between keypoint visual descriptors in images and 3D points in Structure-from-Motion (SfM) models. Traditional image feature extraction techniques, such as SIFT~\cite{lowe2004distinctive}, SURF~\cite{bay2006surf}, ORB~\cite{rublee2011orb}, FREAK~\cite{alahi2012freak, burki2019vizard}, and BRIEF~\cite{calonder2010brief, linegar2015work}, tend to perform poorly under significant changes in viewpoint or illumination. Although learnable features like SuperPoint~\cite{detone2018superpoint}, R2D2~\cite{revaud2019r2d2}, and GIFT~\cite{liu2017gift} have enhanced robustness to some extent, they still face challenges with variations in weather and appearance. Moreover, constructing large-scale 3D maps is highly expensive, and these maps require frequent updates to reflect environmental changes. The cost of storing 3D maps in vehicles further complicates their implementation in city-scale areas.

Recent approaches have transitioned from relying exclusively on appearance information to incorporating semantic data for localization in vectorized maps. These methods generally extract semantic features using pre-trained convolutional neural networks (CNNs) and then link these features through filtering or optimization techniques to achieve precise localization. However, semantic features can be affected by artifacts or missing data, prompting the use of complex handcrafted strategies like graph matching~\cite{wang2021visual} or distance transforms~\cite{pauls2020monocular}. These approaches often require careful tuning of various hyperparameters, making them cumbersome and difficult to generalize.
Notable works, such as BEV-Locator~\cite{zhang2022bev}, focus on centimeter-level localization using HD maps, attempting to directly regress pose offsets with transformer-based models that jointly encode image and map features. Nonetheless, the absence of explicit geometric constraints results in limited accuracy and robustness in direct pose regression. OrienterNet~\cite{sarlin2023orienternet}, designed for relocalization using navigation maps, introduces explicit 3-DoF (Degree-of-Freedom) pose probability modeling after fusing image and map features but still lacks clear modeling of matching relationships in BEV space. MapLocNet~\cite{wu2024maplocnet} enhances OrienterNet's performance with a coarse-to-fine pose regression model but does not address explicit image-map matching.

To tackle the issues present in existing methods, this paper presents an uncertainty-aware, end-to-end visual localization framework that enables adaptive localization across different map formats by integrating matching and search strategies. 
Specifically, drawing inspiration from end-to-end image feature matching~\cite{sun2021loftr,sarlin2020superglue}, we establish both global and local associations between image features and BEV map features to create cross-modal relationships. Our design uniquely incorporates perceptual uncertainty as a guiding mechanism, leveraging multivariate Gaussian for global supervision and contrastive learning to enhance local association robustness. Unlike traditional methods that optimize poses based on matches, we employ a pose decoder to regress pose distributions. This approach constructs refined solution spaces centered around coarse localization uncertainty, enabling the search for precise poses. By integrating these strategies, our method effectively combines the large-scale localization capabilities of matching with the fine-grained precision of registration within localized regions.
Our contributions can be summarized as follows:

\begin{itemize}

\item We introduce a novel uncertainty-aware progressive visual localization framework, supporting various map formats and localization tasks.

\item We propose Perceptual Uncertainty-guided Association, which integrates perception uncertainty into probabilistic modeling, to reduce the impact of low-quality regions.
\item We propose Localization Uncertainty-guided Registration, leveraging coarse pose uncertainty as a prior for fine-grained localization, addressing non-unimodal pose distribution challenges.
\item Extensive experiments on centimeter-level localization in small areas and large-scale relocalization demonstrate that our proposed U-ViLAR achieves state-of-the-art performance across various datasets.
\end{itemize}

%% file: sec/2_related.tex
\section{Related Work}
\subsection{BEV Representation in Visual Perception}
Transforming image features into BEV space can be achieved through geometric or learning-based methods. Geometric approaches, such as Cam2BEV~\cite{cam2bev} and VectorMapNet~\cite{vectormapnet}, apply Inverse Perspective Mapping (IPM) to convert image features into BEV grids. Learning-based methods, such as LSS~\cite{philion2020lift}, predict depth to project image pixels into BEV space, further refined by BEVDet~\cite{huang2021bevdet} and BEVDepth~\cite{bevdepth} for improved 3D perception. Alternative approaches include query-based models like Detr3D~\cite{wang2022detr3d}, which sample image features using 3D queries for regression, and GitNet~\cite{gong2022gitnet}, which introduces a two-stage perspective-to-BEV transformation. Recent methods, such as BEVFormer~\cite{li2022bevformer} and PETR~\cite{liu2022petr}, leverage cross-attention and 3D position encoding to enhance 2D-to-BEV mapping.
Beyond discriminative perception, BEV-centric and world-model representations have also been explored for layout-guided driving video generation, multimodal BEV latent modeling, trajectory-conditioned driving simulation, future scene generation with perception, downstream decision learning, and broader autonomous-driving world-model surveys~\cite{li2024drivingdiffusion,zhang2024bevworld,li2025driverse,liang2025seeing,xiao2025learning,tu2025roleworldmodels}. These advances indicate a broader shift toward structured spatial representations that can bridge visual observations, maps, and planning-oriented scene understanding. Meanwhile, progress in visual generative modeling and 3D scene editing, including focus-driven autoregressive modeling, temporal formulations of image editing, reward-guided video diffusion, drag-based Gaussian editing, and NeRF-based multi-view 3D detection~\cite{li2025fvar,li2025video4edit,yang2025dualdiff+,qu2025drag,huang2025nerf}, further highlights the importance of learning geometry-aware spatial and temporal correspondences. Related advances in visual prompting, visual in-context learning, descriptive caption enhancement, MLLM capability analysis, multimodal memory, and efficient vision feature resampling~\cite{sun2024vrpsam,sun2025exploring,sun2024descriptivecaption,liu2024revisitingmllms,bo2025agenticlearner,feng2025visionremember} also suggest that stronger visual representations can provide more reliable semantic and spatial cues for downstream localization.

\subsection{Visual Localization}
Traditional localization methods rely on geometric feature matching, using hand-crafted descriptors such as SIFT, SURF, and ORB to associate 2D image pixels with 3D scene points~\cite{li2016worldwide, shotton2013scene}. However, these methods are sensitive to viewpoint and illumination variations. Deep learning improves robustness by replacing handcrafted features with learned representations~\cite{detone2018superpoint, revaud2019r2d2}, yet computational costs remain high. Semantic maps provide stable environmental cues such as lane markings and poles, enabling localization through LiDAR-based map matching~\cite{levinson2007map, levinson2010robust} and neural-based pose estimation~\cite{kendall2015posenet}. Further improvements integrate sensor fusion~\cite{suhr2016sensor} and monocular SLAM~\cite{sattler2017large}, enhancing accuracy and robustness. Large-scale localization has shifted towards lightweight navigation maps, where semantic matching between urban images and 2D maps~\cite{automated_map_reading, you_are_here} facilitates positioning. Cross-view approaches~\cite{slicematch, uncertainty_aware_geolocalization} align ground-level and aerial imagery, while OrienterNet~\cite{sarlin2023orienternet} achieves sub-meter precision using BEV-based neural matching with OpenStreetMap~\cite{haklay2008openstreetmap}.
Recent autonomous-driving studies further emphasize spatial reasoning under map or prompt guidance, including marker-based prompt learning for improved spatial understanding and uncertainty-aware visual localization via differentiable association and registration~\cite{zhang2025mpdrive,li2025u}. These works are closely related to our motivation of using explicit BEV-space relationships to obtain robust pose estimates from imperfect observations and maps.

\subsection{End-to-End Localization}
End-to-end (E2E) localization methods directly infer poses from sensor inputs and prior maps, eliminating the need for explicit matching. 
PixLoc~\cite{PixLoc} introduces a differentiable optimization framework, aligning depth features with a reference 3D model to estimate poses in an E2E manner. I2D-Loc~\cite{I2D-Loc} further enhances efficiency by integrating local image-LiDAR depth registration and BPnP-based gradient computation for pose refinement. More recent advancements, such as BEV-Locator~\cite{zhang2022bev}, leverage multi-view images and vectorized global maps, employing a cross-modal Transformer to align semantic map elements with camera images. EgoVM~\cite{egovm} extends this paradigm by achieving centimeter-level localization by integrating point cloud data and lightweight vectorized maps. However, although these methods can output localization results end-to-end, they tend to be relatively simplistic and lack fine precision. We designed the model structure based on priors from the localization domain while ensuring gradient continuity, achieving performance that far surpasses previous work.

\begin{figure*}
\begin{center}
\includegraphics[width=0.95\linewidth]{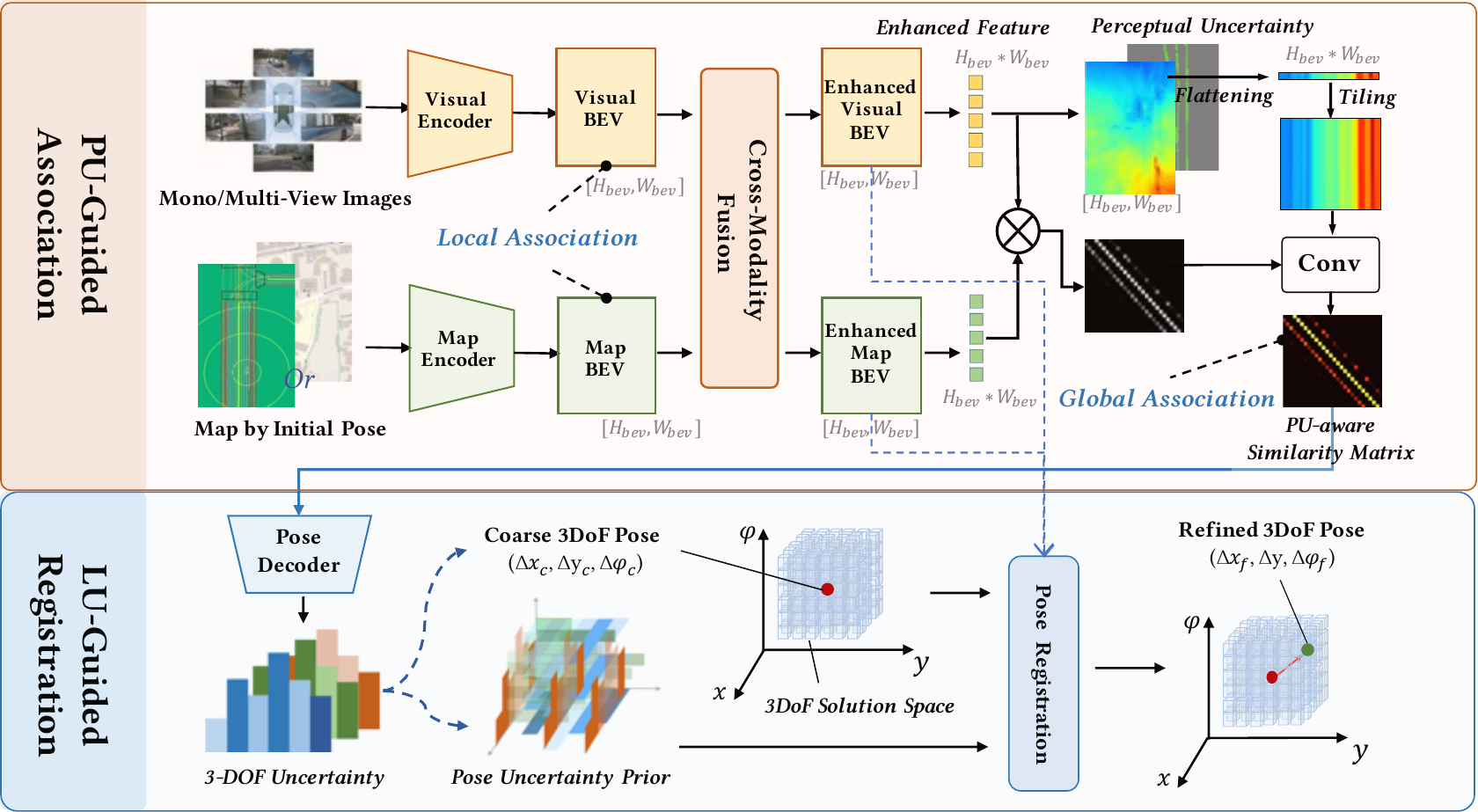}
\vspace{-4mm}
\end{center}
   \caption{\textbf{An overview of the proposed U-ViLAR. } First, we extract features from image and map inputs, then align these features into BEV space to obtain visual and map features. These features undergo cross-modal fusion to enhance visual and map features (Sec.~\ref{sec:m1}). During the Perceptual Uncertainty-guided (PU-Guided) Association phase (Sec.~\ref{sec:m2}), we construct a similarity matrix from the features above and refine it through probabilistic guidance using perceptual uncertainty derived from visual features, yielding a Perceptual Uncertainty-aware (PU-aware) similarity matrix. During the Localization Uncertainty-guided (LU-Guided) Registration phase, this matrix is processed by the pose decoder to yield a coarse pose distribution and a coarse pose. These outputs serve as the pose uncertainty prior and the center of the solution space for Pose Registration. Finally, precise poses are obtained by registering the enhanced visual-map BEV (Sec.~\ref{sec:m3}).}
\vspace{-4mm}
\label{fig:m1}
\end{figure*}

%% file: sec/3_method.tex
\section{Method}
\noindent\textbf{Problem Definition.} 
Given an initial estimate $\mathbf{P}_0 = (x_0, y_0, \varphi_0)^\top \in \mathbb{R}^3$ obtained from noisy sensor measurements, our objective is to predict a minimal correction $\Delta\mathbf{P} = (\delta x, \delta y, \delta\varphi)$ that aligns this estimate with the ground-truth pose $\mathbf{P}_{\mathrm{gt}} \in \mathbb{R}^3$. This correction is expressed as a rigid-body transformation in the Special Euclidean group $\mathrm{SE}(2)$, parameterized by the transformation matrix $\mathbf{T} \in \mathbb{R}^{3 \times 3}$:
\[
\mathbf{T} = 
\begin{bmatrix} 
\cos\Delta\varphi & -\sin\Delta\varphi & \Delta x \\ 
\sin\Delta\varphi & \cos\Delta\varphi & \Delta y \\ 
0 & 0 & 1 
\end{bmatrix}.
\]
To ensure accurate pose correction, we minimize the squared Euclidean alignment error:
$$ \min \mathcal{L} = \left\| \mathbf{T} \mathbf{P}_0 - \mathbf{P}_{\mathrm{gt}} \right\|_2^2, $$
where $\mathbf{T}$ is inferred by the end-to-end differentiable network.
localization accuracy.

\vspace{2pt} \noindent\textbf{Overview.} 
As depicted in Fig.~\ref{fig:m1}, U-ViLAR comprises two key components: Perceptual Uncertainty-guided (PU-Guided) Association and Localization Uncertainty-guided (LU-Guided) Registration. The LU-Guided Registration component enhances the localization output by utilizing the coarse pose and distribution generated by the PU-Guided Association.

\subsection{BEV Feature Extraction and Fusion}\label{sec:m1}
As shown in the upper part of Fig.~\ref{fig:m1}, to establish correspondences between images and maps for accurate localization, we first extract features from both image and map inputs, and align them to the BEV space. Then, feature fusion is performed within this unified representation.

\noindent\textbf{Visual Encoder.}  
For monocular or multi-view images $\{I_i|i=1,2,\cdots,N\}$, we feed them into a shared backbone network (e.g., ResNet~\cite{He2016DeepRL}) to obtain $\{F^v_i|i=1,2,\cdots,N\}$, where $F^v_i\in \mathbb{R}^{H_v\times W_v \times C_v}$ represents the feature map of the $i$-th view, with $H_v$ and $W_v$ denoting the height and width of the extracted features, respectively. Then, following the approach of BEVFormer~\cite{li2022bevformer}, we apply a view transformation to the extracted visual features and project them into the BEV space, resulting in the visual BEV features ${\textbf{F}}^{BEV}_v \in {\mathbb{R} ^ {H^{BEV}_v \times W^{BEV}_v \times C^{BEV}_v}}$, where the visual BEV space size is ${{H}^{BEV}_v \times {W}^{BEV}_v}$.

\noindent\textbf{Map Encoder. }
Our method supports HD maps or navigation maps as input, with similar map processing strategies adopted in previous research, such as \cite{zhang2022bev, sarlin2023orienternet}. When employing HD maps, we utilize the high-definition map data provided by the respective datasets. For navigation maps, OpenStreetMap~\cite{haklay2008openstreetmap} (OSM) serves as the data source.
Various elements (lanes, curbs, etc.) are rasterized with a fixed sampling distance $\Delta$ (e.g., 25 cm/pixel), resulting in $I_{\text{map}} \in \mathbb{R} ^{H_m \times W_m \times N_m}$, where $H_m \times W_m$ represents the rasterized result of a particular element, and $N_m$ is the number of elements in the map. The rasterized map image $\mathbf{I}_{\text{map}}$ is then passed through a U-Net-like backbone~\cite{ronneberger2015u} to extract the map features ${\textbf{F}}^{BEV}_m \in {\mathbb{R} ^ {{H}^{BEV}_m \times {W}^{BEV}_m \times {C}^{BEV}_m}}$, where the map BEV space size is ${{H}^{BEV}_m \times {W}^{BEV}_m}$.

\noindent\textbf{Cross-Modality Fusion. }
We perform cross-modal fusion by alternately applying self-attention (SA) and cross-attention (CA) to integrate visual BEV feature $\mathbf{F}^{\text{BEV}}_v$ and map BEV feature $\mathbf{F}^{\text{BEV}}_m$, resulting in the enhanced features $\mathbf{F}_v^{\prime\text{BEV}}$ and $\mathbf{F}_m^{\prime\text{BEV}}$:
\begin{align}
    \mathbf{F}_v^{\prime\text{BEV}} &= \operatorname{CA}(\operatorname{SA}(\mathbf{F}^{\text{BEV}}_v), \operatorname{SA}(\mathbf{F}^{\text{BEV}}_m)), \\
    \mathbf{F}_m^{\prime\text{BEV}} &= \operatorname{CA}(\operatorname{SA}(\mathbf{F}^{\text{BEV}}_m), \operatorname{SA}(\mathbf{F}^{\text{BEV}}_v)),
\end{align}
where $\mathrm{SA}(\cdot)$ and $\mathrm{CA}(Q,K)$ represent self-attention and cross-attention with query-key pairs.


\begin{figure}[t]
\centering
\includegraphics[width=\linewidth]{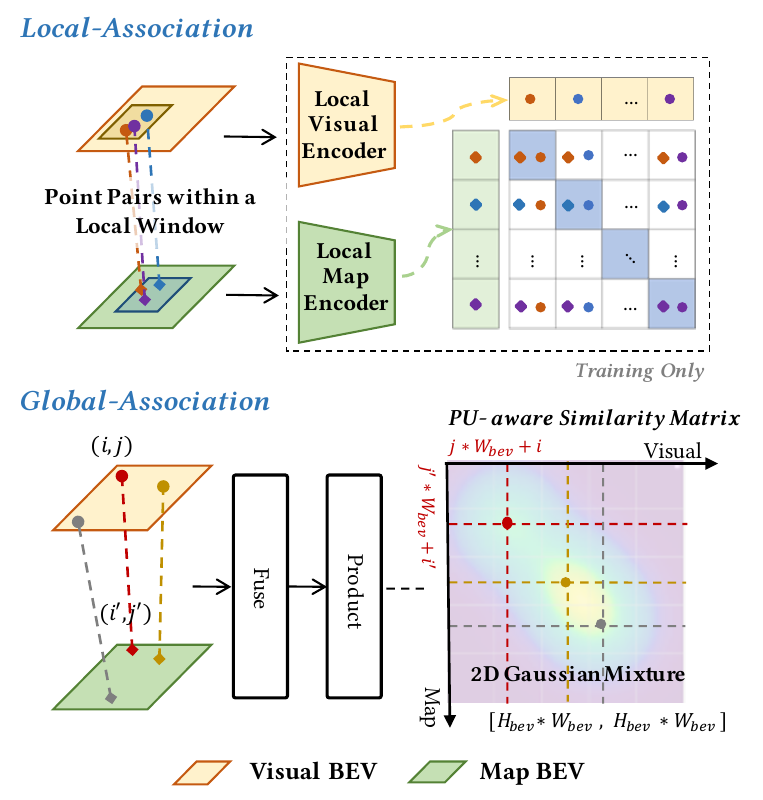}
\caption{\textbf{Illustration of Local Association and Global Association. } The BEV features in the upper and lower parts are the original and the enhanced ones, respectively. }
\centering
\vspace{-3mm}
\label{fig:m2}
\end{figure}

\subsection{Perceptual Uncertainty-guided Association}\label{sec:m2}
\noindent\textbf{Overview. }When associating semantic elements between vision and maps, visual inputs often suffer from perceptual degradation, such as occlusions of key semantic elements. To tackle this issue, we propose a method that models perceptual uncertainty to guide the cross-modal association between visual features and map features in BEV space. Specifically, we supervise the perceptual uncertainty of $\mathbf{F}_v^{\prime\text{BEV}}$, which supports perception tasks and facilitates the generation of BEV perceptual uncertainty, as illustrated in Fig.~\ref{fig:m1}. We then incorporate this perceptual uncertainty into the association modeling between $\mathbf{F}_v^{\prime\text{BEV}}$ and $\mathbf{F}_m^{\prime\text{BEV}}$, creating a perceptual uncertainty-aware similarity matrix. Furthermore, to enhance the model's ability to establish fine-grained cross-modal associations for spatially close points within local regions, we sample local windows from $\mathbf{F}^{\text{BEV}}_v$ and $\mathbf{F}^{\text{BEV}}_m$, using contrastive learning to supervise the associations within these sampled windows.

\noindent\textbf{Perceptual Uncertainty. }
We enhance the feature learning by incorporating road structure-aware supervision with uncertainty prediction on $\mathbf{F}_v^{\prime\text{BEV}}$. Inspired by~\cite{ref3}, we estimate a pixel-wise uncertainty field $\mathbf{U_P} \in \mathbb{R}^{H_v^{\text{BEV}} \times W_v^{\text{BEV}}}$ to guide the feature refinement. The uncertainty loss is formulated as:
\begin{equation}
    \mathcal{L}_{p} = \exp(-\mathbf{U_P}) \| \mathbf{F} - \mathbf{\hat{F}} \|_1 + 2\mathbf{U_P},
\end{equation}
where $\mathbf{\hat{F}}$ denotes the BEV feature estimate refined by road structure segmentation prediction, and $\mathbf{U_P}$ models the heteroscedasticity in feature prediction. This loss dynamically adjusts reconstruction weights through the $\exp(-\mathbf{U_P})$ term, relaxing constraints in high-uncertainty regions (e.g., occlusion boundaries) while maintaining strict feature alignment in confident areas.

\noindent\textbf{Global Association. }
By computing the similarity matrix $\mathcal{S} \in \mathbb{R}^{(H_v^{\text{BEV}} W_v^{\text{BEV}}) \times (H_m^{\text{BEV}} W_m^{\text{BEV}})}$ between the visual BEV feature map $\mathbf{F}_v^{\text{BEV}}$ and the map BEV feature map $\mathbf{F}_m^{\text{BEV}}$, we model the cross-modal similarity relationship:  
\begin{equation}
\mathcal{S}(i, j) = \left\langle \mathbf{F}_v^{\text{BEV}}(i), \mathbf{F}_m^{\text{BEV}}(j) \right\rangle,
\end{equation}
where $i \in \{1,...,H_v^{\text{BEV}} W_v^{\text{BEV}}\}$ and $j \in \{1,...,H_m^{\text{BEV}} W_m^{\text{BEV}}\}$ represent the spatial unit indices in the visual and map BEV, respectively.

To incorporate perceptual uncertainty, we begin by flattening the 2D uncertainty mask $\mathbf{U_P} \in \mathbb{R}^{H_{\text{BEV}} \times W_{\text{BEV}}}$ into a 1D vector $\mathbf{U_P}^{\text{flat}} \in \mathbb{R}^{H_v^{\text{BEV}} W_v^{\text{BEV}}}$. This flattened vector is then tiled to align with the spatial dimensions of $\mathcal{S}$, resulting in a tiled matrix $\mathbf{U_P}^{\text{tile}} \in \mathbb{R}^{(H_v^{\text{BEV}} W_v^{\text{BEV}}) \times (H_m^{\text{BEV}} W_m^{\text{BEV}})}$. 

Subsequently, we perform a channel-wise concatenation of the original similarity matrix \(\mathcal{S}\) with the tiled uncertainty matrix \(\mathbf{U_P}^{\text{tile}}\). This concatenated result is then processed through a CNN, which consists of a lightweight network with a 1×1 convolutional layer and ReLU activation, ultimately generating the perceptual uncertainty-aware similarity matrix \(\mathcal{S}_{\text{uncert}}\). 
The final association distribution $\mathbf{P} \in \mathbb{R}^{(H_v^{\text{BEV}} W_v^{\text{BEV}}) \times (H_m^{\text{BEV}} W_m^{\text{BEV}})}$ is computed by applying Softmax normalization to $\mathcal{S}_{\text{uncert}}$:  
\begin{equation}
\mathbf{P}_{i,j} = \frac{\exp\left(\mathcal{S}_{\text{uncert}}(i,j)\right)}{\sum_{k=1}^{H_m^{\text{BEV}} W_m^{\text{BEV}}} \exp\left(\mathcal{S}_{\text{uncert}}(i,k)\right)}.
\end{equation}

As shown in Fig.~\ref{fig:m2}, to avoid quantization errors caused by supervision with a hard binary mask, we construct a similarity soft supervision matrix $\mathbf{G} \in \mathbb{R}^{(H_v^{\text{BEV}} W_v^{\text{BEV}}) \times (H_m^{\text{BEV}} W_m^{\text{BEV}})}$ using a multivariate Gaussian distribution:  
\begin{equation}
\mathbf{G}_{i,j} = 
\begin{cases}
\exp\left(-\frac{d^2(i,j)}{2\sigma_i^2}\right) & d(i,j) \leq 3\sigma_i \\
0 & \text{otherwise},
\end{cases}
\end{equation}
where $d(i,j)$ represents the geometric offset between the two spatial units, and $\sigma_i$ is an adaptive smoothing factor.  
For each visual unit \( i \), normalization is performed along the map dimension to obtain \(\tilde{\mathbf{G}}_{i,j}\).
The supervision loss is defined as the cross-entropy between the predicted association distribution $\mathbf{P}$ and the soft target $\tilde{\mathbf{G}}$:  
\begin{equation}
\mathcal{L}_{\text{GM}} = -\sum_{i=1}^{H_v^{\text{BEV}} W_v^{\text{BEV}}} \sum_{j=1}^{H_m^{\text{BEV}} W_m^{\text{BEV}}} \tilde{\mathbf{G}}_{i,j} \log \mathbf{P}_{i,j}.
\end{equation}

\noindent\textbf{Local Association. }
Initially, we leverage the pose ground truth to create spatial correspondences between the visual BEV features, \(\mathbf{F}^{\text{BEV}}_v\), and the map BEV features, \(\mathbf{F}^{\text{BEV}}_m\). Following this, \(\mathbf{M}\) anchor points are uniformly distributed across the space, with each anchor overseeing a local window of dimensions \([\mathbf{H}_p, \mathbf{W}_p]\). For each of these anchors, \(\mathbf{K}\) pairs of ground-truth points are sampled, with features $\mathbf{F}^{\text{map}}_i$ and $\mathbf{F}^{\text{vis}}_i$ extracted by train-only encoders.
Subsequently, a symmetric local similarity matrix $\mathbf{S} \in \mathbb{R}^{\mathbf{K} \times \mathbf{K}}$ is constructed in the BEV local space using contrastive learning principles, where elements are computed as:
\begin{equation}
    \mathbf{S}_{ij} = \frac{\mathbf{F}^{\text{vis}}_i \cdot \mathbf{F}^{\text{map}}_j}{\|\mathbf{F}^{\text{vis}}_i\| \|\mathbf{F}^{\text{map}}_j\|}, \quad \forall i,j \in \{1,\ldots,\mathbf{K}\}.
\end{equation}

The similarity of positive sample pairs is optimized via a symmetric cross-entropy loss:
\begin{equation}
    \mathcal{L}_{\text{LM}} = \frac{1}{2\mathbf{K}} \sum_{i=1}^{\mathbf{K}} \left[ -\log \frac{e^{\mathbf{S}_{ii}}}{\sum_{j=1}^{\mathbf{K}} e^{\mathbf{S}_{ij}}} -\log \frac{e^{\mathbf{S}_{ii}}}{\sum_{j=1}^{\mathbf{K}} e^{\mathbf{S}_{ji}}} \right].
\end{equation}


\begin{figure}[t]
\centering
\includegraphics[width=\linewidth]{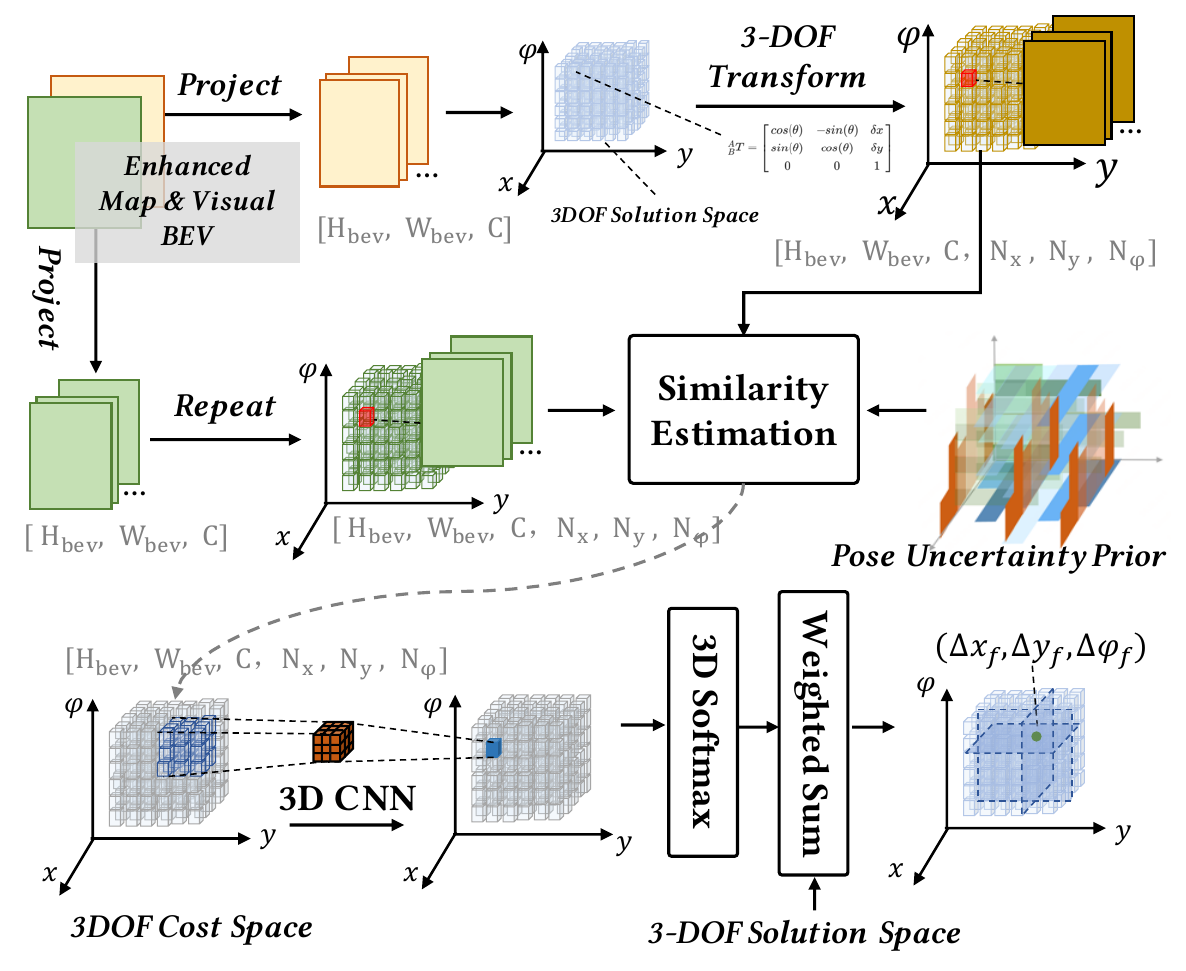}
\caption{\textbf{Illustration of the Pose Registration. } The solution space is constructed centered around the coarse pose, and the precise pose is obtained by estimating the probability of each candidate pose by the coarse prior.}
\centering
\label{fig:m3}
\end{figure}

\subsection{Localization Uncertainty-guided Registration}\label{sec:m3}
\noindent\textbf{Overview. }
Most second-stage coarse-to-fine localization approaches rely heavily on the initial coarse pose input, yielding refined results that remain closely tied to the initial estimation. This oversight neglects potential degeneracy in specific degrees of freedom or multimodal distributions, where poor-quality coarse poses would render subsequent local regression meaningless. To address this, we construct the pose solution space not only from the coarse pose but also model a 3D joint probability distribution from the coarse 3DoF pose estimates, enabling error correction through probabilistic-guided pose registration and weighted fusion.

\noindent\textbf{Localization Uncertainty. }
As shown in the lower part of Fig.~\ref{fig:m1}, we regress 3DoF probability distributions via a pose decoder from the perceptual uncertainty-aware similarity matrix $\mathcal{S}_{\text{uncert}}$. Specifically, we discretize each DoF: $x \in [x_{\min}^c, x_{\max}^c]$ into $N_x$ bins, $y \in [y_{\min}^c, y_{\max}^c]$ into $N_y$ bins, and orientation $\varphi \in [\varphi_{\min}^c, \varphi_{\max}^c]$ into $N_\varphi$ bins. The decoder outputs independent probability distributions $\mathbf{p}_x \in \mathbb{R}^{N_x}$, $\mathbf{p}_y \in \mathbb{R}^{N_y}$, and $\mathbf{p}_\varphi \in \mathbb{R}^{N_\varphi}$ (with $\sum \mathbf{p}_* = 1$). Supervision is provided by Gaussian distributions centered at ground truth with standard deviations controlling distribution widths, normalized to form probability targets.

The uncertainty for each DoF (lateral $x$, longitudinal $y$, and rotational $\varphi$) is quantified via Shannon entropy:
\begin{equation}
\mathbf{U}_d = -\sum_{n=1}^{N_d} \mathbf{p}_d^{(n)}\log\mathbf{p}_d^{(n)},\quad d \in \{x,y,\varphi\}
\end{equation}
where $N_d$ denotes bin count for DoF $d$, and $\mathbf{p}_d^{(n)}$ represents the $n$-th bin probability. This generalized entropy formulation preserves DoF-specific parameters (e.g., varying $N_x,N_y,N_\varphi$) while maintaining physical consistency with individual uncertainty metrics.

\noindent\textbf{Coarse Pose Regression and 3D Joint Distribution. }
The 3DoF distributions are concatenated and fed to an MLP to regress coarse pose estimates $(x_c, y_c, \varphi_c)$. Simultaneously, we construct a joint probability space via Cartesian product:
\begin{equation}
\mathbf{P}_{\text{uncert}}(i,j,k) = \mathbf{p}_x^{(i)} \cdot \mathbf{p}_y^{(j)} \cdot \mathbf{p}_\varphi^{(k)},\quad \mathbf{P}_{\text{uncert}} \in \mathbb{R}^{N_x \times N_y \times N_\varphi}.
\end{equation}

\noindent\textbf{Localization Uncertainty-Guided Pose Registration. }
As shown in Fig.~\ref{fig:m3}, based on the coarse pose $(x_c, y_c, \varphi_c)$ and joint prior $\mathbf{P}_{\text{uncert}}$, we define a 3D solution space $\Omega$ centered around the coarse pose, with ranges $[x_c \pm \Delta_x]$, $[y_c \pm \Delta_y]$, and $[\varphi_c \pm \Delta_\varphi]$. This space is discretized into $(H_s, W_s, D_s)$ candidates, indexed by $(i, j, k)$, with resolutions $(\delta_x, \delta_y, \delta_\varphi) = (\frac{2\Delta_x}{H_s}, \frac{2\Delta_y}{W_s}, \frac{2\Delta_\varphi}{D_s})$.
Consequently, visual features $\mathbf{F}_v^{\prime\text{BEV}}$ undergo 3DoF geometric transformations in solution space $\Omega$:
\begin{equation}
\begin{aligned}
\mathbf{F}_v^T(h,w,c,i,j,k) &= \mathcal{T}_{\theta_{i,j,k}}(\mathbf{F}_v^{\text{BEV}}(h,w,c)), \\
\theta_{i,j,k} = (x_c - \Delta_x + i\delta_x,\ y_c &- \Delta_y + j\delta_y,\ \varphi_c - \Delta_\varphi + k\delta_\varphi),
\end{aligned}
\end{equation}
where $\mathcal{T}_\theta(\cdot)$ denotes the 3DoF transform operator. Map features $\mathbf{F}_m^{\prime\text{BEV}}$ are aligned and broadcast to match dimensions as $\mathbf{F}_m^B$. The feature difference tensor $\mathbf{D}_{\text{cost}}$ is computed as the L2 norm between the transformed visual features $\mathbf{F}_v^T$ and the aligned map features $\mathbf{F}_m^B$.

Subsequently, the joint probability $\mathbf{P}_{\text{uncert}}$ is fused with the feature difference tensor $\mathbf{D}_{\text{cost}}$ via gating, serving as a prior for the pose registration:
\begin{equation}
\mathbf{D}_{\text{fused}} = \mathbf{D}_{\text{cost}} + \lambda \cdot \text{repeat}(\text{unsqueeze}(\mathbf{P}_{\text{uncert}})). 
\end{equation}
where $\lambda$ is a learnable parameter. After convolutional processing of $\mathbf{D}_{\text{fused}}$, the cost tensor in cost space tensor $\mathbf{C} \in \mathbb{R}^{H_s \times W_s \times D_s}$ is obtained, and the refined pose $(x_f, y_f, \varphi_f)$ is obtained via softmax-weighted fusion:
\begin{equation}
\begin{aligned}
(x_f, y_f, \varphi_f) = \sum_{i,j,k} \sigma(\mathbf{C} &+ \gamma \mathbf{P}_{\text{uncert}})_{i,j,k} 
\cdot (x_c-\Delta_x+i\delta_x, \\ 
y_c-&\Delta_y+j\delta_y, \varphi_c-\Delta_\varphi+k\delta_\varphi),
\end{aligned}
\end{equation}
where $\gamma$ controls prior strength and $\sigma(\cdot)$ denotes 3D softmax. The parameter $\gamma$ dynamically modulates the balance of weights between the probabilistic coarse pose prior and data-driven signals during pose refinement, enabling the model to adaptively adjust its reliance on prior knowledge according to the uncertainty inherent in the input scene.

\begin{table*}[t]
\centering
\footnotesize{\input{tables/ori1}}
\caption{\textbf{Centimeter-level localization results on nuScenes and SRoad using their respective HD maps as input.} 
We use \textbf{bold} font to highlight the best results. S and M represent monocular and multi-view inputs, respectively.}%
\label{tab:ori1}
\end{table*}

\begin{table*}[t]
\centering
\footnotesize{\input{tables/o2-kitti}}
\caption{\textbf{Localization results on KITTI and SRoad dataset using OSM as the map input.}
We use \textbf{bold} font to highlight the best results. S and M represent monocular and multi-view inputs, respectively.
}
\vspace{-4mm}
\label{tab:o2-kitti}
\end{table*}

%% file: tables/ori1.tex
\begin{tabular}{lccccccc}
\toprule
\multirow{2}{*}[-.3em]{Methods} & 
\multirow{2}{*}[-.3em]{\makecell{Inputs}} 
& \multicolumn{2}{c}{Lateral Error $\downarrow$}
& \multicolumn{2}{c}{Longitudinal Error $\downarrow$}
& \multicolumn{2}{c}{Orientation Error $\downarrow$}\\ 
\cmidrule(lr){3-4}
\cmidrule(lr){5-6}
\cmidrule(lr){7-8}
& & MAE(m) & RMSE(m) &  MAE(m) & RMSE(m) & MAE(\si{\degree}) & RMSE(\si{\degree})\\ 
\midrule
BEV-Locator-M & nuScenes + HD map  &  0.076 & - & 0.178 & - & 0.510 & - \\
Ours-M & nuScenes + HD map  &  \textbf{0.040}&  0.049& \textbf{0.140}&  0.158& \textbf{0.075}&  0.089\\ 

\midrule

OrienterNet-S & nuScenes + HD map  &  0.079& 0.154& 0.279& 0.473& 0.114& 0.143 \\
Ours-S & nuScenes + HD map  & \textbf{0.069 }&  \textbf{0.134} & \textbf{0.223} &  \textbf{0.324} & \textbf{0.092} &  \textbf{0.105}  \\

\midrule

ICP-based-M& SRoad + HD map  & 0.204 & 0.320 & 0.614 & 1.106 & 0.186 & 0.204\\
Ours-M& SRoad + HD map  & \textbf{0.110}& \textbf{0.136} & \textbf{0.284} & \textbf{0.322}& \textbf{0.090} & \textbf{0.124}\\

\midrule

OrienterNet-S & SRoad + HD map  & 0.186& 0.241& 0.492& 0.724& 0.160& 0.245\\
Ours-S & SRoad + HD map  & \textbf{0.165}& \textbf{0.193}& \textbf{0.390}& \textbf{0.484}& \textbf{0.122}& \textbf{0.154}\\

\bottomrule
\end{tabular}

%% file: tables/o2-kitti.tex
\begin{tabular}{lcccccccccc}
\toprule
\multirow{2}{*}[-.3em]{Methods} & \multirow{2}{*}[-.3em]{\makecell{Inputs}}
& \multicolumn{3}{c}{Lateral Recall@Xm $\uparrow$}
& \multicolumn{3}{c}{Longitudinal Recall@Xm $\uparrow$}
& \multicolumn{3}{c}{Orientation Recall@X\si{\degree} $\uparrow$}\\ 
\cmidrule(lr){3-5}
\cmidrule(lr){6-8}
\cmidrule(lr){9-11}

& & 1m & 3m & 5m  & 1m & 3m & 5m & 1\si{\degree} & 3\si{\degree} & 5\si{\degree}\\ 

\midrule

OrienterNet-S  & KITTI + OSM & 51.26 & 84.77 & 91.81  & 22.39 & 46.79 & 57.81   & 20.41 & 52.24 & 73.53  \\
Ours-S  & KITTI + OSM & \textbf{69.12} & \textbf{91.25} & \textbf{93.68}  & \textbf{32.04} & \textbf{63.00} & \textbf{70.20} & \textbf{64.92} & \textbf{94.84} & \textbf{97.44}  \\

\midrule
OrienterNet-S & SRoad + OSM & 35.23 & 76.38 & 84.51 & 15.20 & 35.40 & 48.91 & 16.21 & 40.12 & 62.77 \\
Ours-S  & SRoad + OSM & \textbf{52.90} & \textbf{86.20} & \textbf{88.03}  & \textbf{27.94} & \textbf{45.53} & \textbf{55.84} & \textbf{49.91} & \textbf{82.50} & \textbf{86.01} \\ 

\bottomrule
\end{tabular}

%% file: sec/4_exp.tex
\section{Experiment}

\subsection{Experimental Setup}
\noindent\textbf{Datasets. }
We evaluate our method on three datasets: nuScenes~\cite{caesar2020nuscenes}, a widely used autonomous driving dataset containing 1,000 scenarios with over 28,000 frames for training and 6,000 frames for validation; KITTI~\cite{geiger2013vision}, which includes 39.2 km of visual odometry sequences; and our self-collected SRoad dataset, which contains over 500,000 frames and features more complex road structures: over 60\% of the scenarios present specific challenges, such as intersections, merging/diverging zones, congested areas, or areas under viaducts (details in {\textbf{Appendix}}).

\begin{table}[t]
\centering
\footnotesize{\input{tables/o2-nus}}
\vspace{-2mm}
\caption{
\textbf{Localization results on the nuScenes dataset using OSM as the map input.}
We compared our method with existing approaches based on both mono-view and multi-view. Similarly, we use \textbf{bold} font to highlight the best results. S and M represent monocular and multi-view inputs, respectively. }
\vspace{-6mm}
\label{tab:o2-nus}
\end{table}

\noindent\textbf{Tasks, Metrics and Comparison Methods.}
We comprehensively compared previous studies across various datasets and map sources. The localization tasks are divided into two categories:
\textit{\textbf{(a) Fine-grained Localization:} }
We conducted experiments using the nuScenes and SRoad datasets, utilizing their respective HD maps, and employed an ICP method as the rule-based baseline (details in {\textbf{Appendix}}). Our method was compared with BEV-Locator and OrienterNet, using MAE and RMSE to evaluate localization performance.
\textit{\textbf{(b) Large-scale Relocalization:} }
We used nuScenes, KITTI, and SRoad datasets with OSM map input. Following Shi et al.~\cite{shi2022beyond}, we computed recall rates for Lateral, Longitudinal, and Orientation at fixed thresholds. On nuScenes, we evaluated recall for position and orientation to align with metrics of previous methods.

\noindent\textbf{Implementation Details. } 
Aligned with previous work, in fine-grained localization, small random transformations (rotation $\theta \in [-2^\circ, 2^\circ]$, translation $t \in [-2\text{m}, 2\text{m}]$) are applied to HD maps to simulate GPS noise, followed by extracting a 120m $\times$ 120m search region centered on the ego vehicle. For relocalization, larger perturbations (rotation $\theta \in [-30^\circ, 30^\circ]$, translation $t \in [-30\text{m}, 30\text{m}]$) are introduced to address significant pose deviations, processing a 128m $\times$ 128m search area on the navigation map (details in {\textbf{Appendix}}).


\subsection{Results}
\noindent\textbf{Fine-grained Localization. }
\Cref{tab:ori1} presents the experimental results of centimeter-level localization tasks using different datasets and their corresponding HD maps. The MAE and RMSE metrics demonstrate the superiority of our approach. Notably, on the nuScenes dataset, we significantly reduced horizontal and vertical MAE compared to BEV-Locator, with an 85.3\% decrease in Orientation MAE ($0.075$ \textit{v.s.} $0.510$). Compared to the monocular version of OrienterNet, our errors were comprehensively reduced, particularly the difference in RMSE, reflecting the stability of our localization method. Our method shows significant advantages over traditional ICP-based industrial standard methods for the challenging SRoad dataset.

\noindent\textbf{Large-scale Relocalization. }
As shown in \Cref{tab:o2-kitti} and \Cref{tab:o2-nus}, we compared our method with previous approaches on the KITTI, nuScenes, and SRoad datasets to evaluate large-scale relocalization capabilities using OSM as the map input. Our method consistently outperforms existing approaches when the threshold is set to 1m, 3m, and 5m, demonstrating its robust capability in both coarse relocation and fine-grained localization.

\noindent\textbf{Runtime Analysis. }
We deploy our model on an NVIDIA V100 GPU, achieving an impressive 28 frames per second (FPS). This high efficiency stems from our training-only optimizations, which enhance model performance without introducing additional latency. 
On NVIDIA Orin, a widely used autonomous driving platform, we apply TensorRT INT8 quantization to the BEV encoder, optimizing efficiency and achieving a 15 FPS runtime for the entire model. 


\subsection{Ablation Study}
To assess the effectiveness of each component of our method, we performed a series of ablation experiments using the nuScenes dataset along with its corresponding HD map. The experimental results on other datasets and our experiments using a Kalman filter to validate the value of localization uncertainty output further are provided in {\textbf{Appendix.}}

\noindent\textbf{PU-Guided Association and LU-Guided Registration. }
We conducted experiments using multi-view images as shown in \Cref{tab:a1}. Comparing S2 and S3 with S1 shows that PU-Guided Association significantly impacts localization more than LU-Guided Registration. Removing all components (S4) results in severe performance degradation. These findings underscore the necessity of combining PU-Guided Association and LU-Guided Registration to optimize localization.

\noindent\textbf{Local Association, Global Association and Perceptual Uncertainty. }
As shown in \Cref{tab:a2}, when Perceptual Uncertainty is removed and the original Similarity Matrix is used instead of the PU-aware Similarity Matrix, a significant increase in localization error is observed, underscoring the crucial role of Perceptual Uncertainty in the process. Conversely, Local Association has a relatively small impact on localization accuracy.

\noindent\textbf{Pose Uncertainty Prior. }
Based on the comparison of the two rows in \Cref{tab:a3}, it is clear that omitting the Pose Uncertainty Prior as input information for Registration results in some deterioration of our primary metric, MAE, reflecting a decline in localization performance. Notably, RMSE experiences an even more pronounced degradation. This aligns with the fundamental design purpose of the Pose Uncertainty Prior, which is intended to be particularly effective in the less common non-unimodal scenarios within the dataset.

\begin{table}[t]
\centering
\footnotesize{\input{tables/a1}}
\caption{\textbf{Ablation experiments on the key components of our method using the nuScenes dataset and HD map.} `Assoc.' and `Reg.' denote PU-Guided Association and LU-Guided Registration, respectively. `Lat.', `Lon.', and `Ori.' stand for Lateral, Longitudinal, and Orientation, respectively.}
\label{tab:a1}
\end{table}
\begin{table}[t]
\centering
\footnotesize{\input{tables/a2}}
\caption{\textbf{Ablation experiments of the key components of PU-Guided Association on the nuScenes dataset and HD map.} `LA.', `GA.', `PU.' represent Local Association, Global Association, and Perceptual Uncertainty, respectively.}
\label{tab:a2}
\end{table}
\begin{table}[t]
\centering
\footnotesize{\input{tables/a3}}
\caption{\textbf{Ablation results of pose distribution prior of LU-Guided Registration on the nuScenes dataset and HD map.} `PDP.' represent Pose Distribution Prior and `M.', `R.' represent MAE and RMSE, respectively.}
\label{tab:a3}
\end{table}

\begin{figure}[!h]
\centering
\vspace{-2mm}
\includegraphics[width=\linewidth]{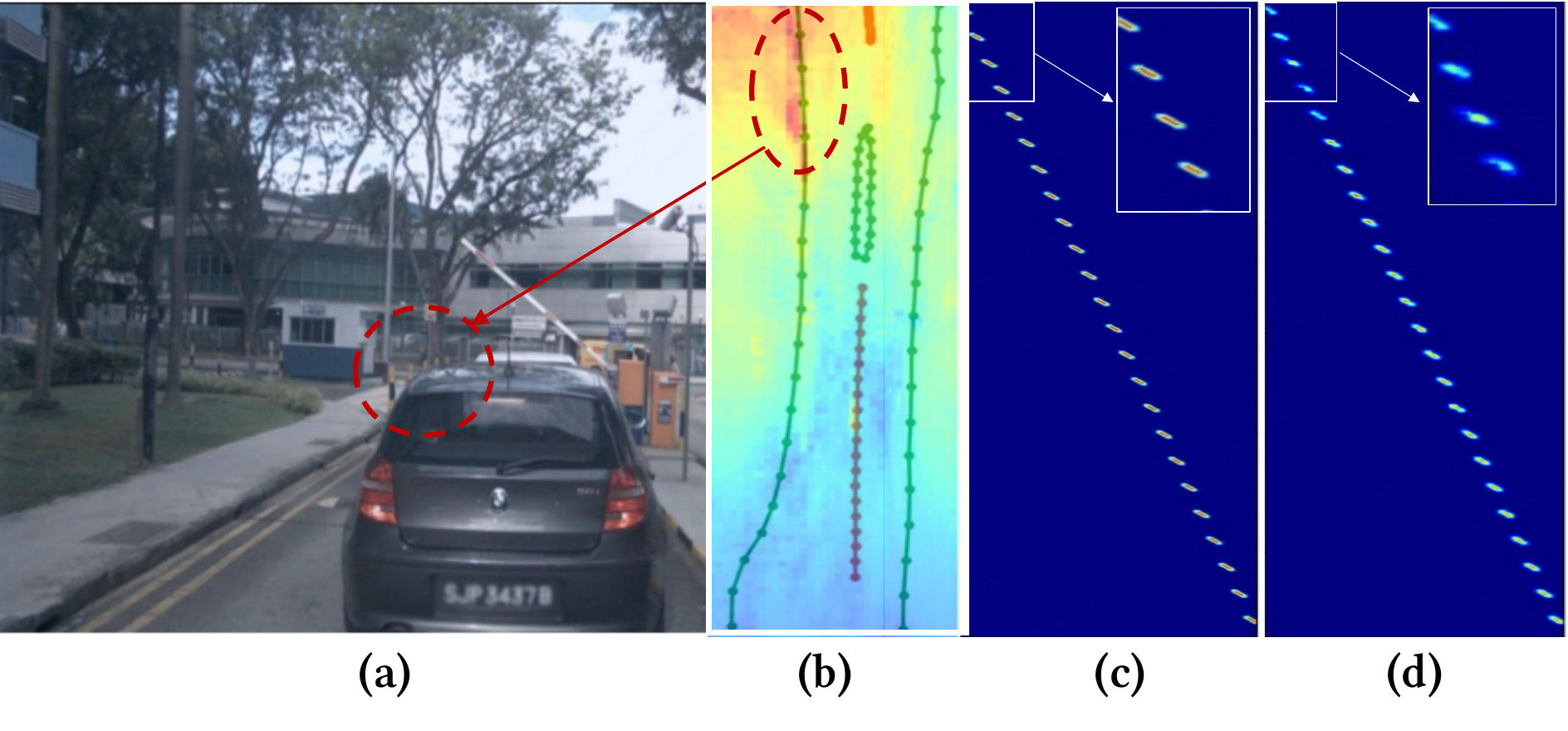}
\caption{\textbf{Visualization on the nuScenes validation set.} From left to right: (a) Front view, (b) Road-structure predictions overlaid with a BEV uncertainty heatmap, where red circles highlight regions of high uncertainty, (c) Ground truth and (d) Inference results of the PU-aware Similarity Matrix.}
\centering
\vspace{-3mm}
\label{fig:e1}
\end{figure}

\begin{figure}[!h]
\centering
\vspace{-2mm}
\includegraphics[width=\linewidth]{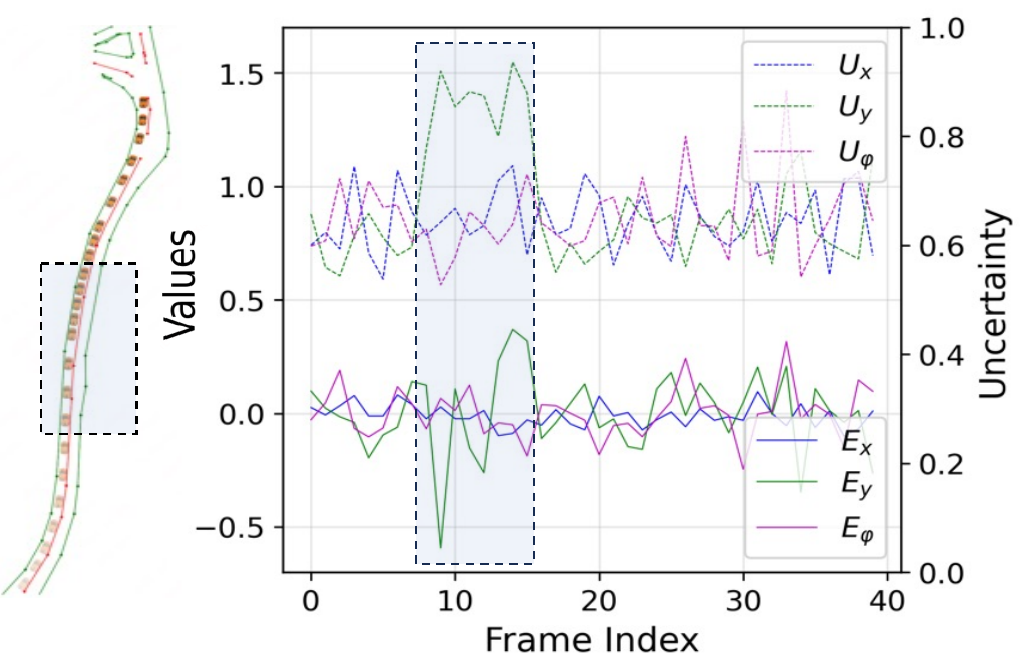}
\caption{\textbf{Trajectory segment from nuScenes validation set (left) and per-frame localization results (right).} $E_x$, $E_y$, $E_\varphi$: lateral, longitudinal, orientation errors; $U_x$, $U_y$, $U_\varphi$: lateral, longitudinal, orientation uncertainties.}
\centering
\vspace{-2mm}
\label{fig:e2}
\end{figure}

\subsection{Qualitative Analysis}

\noindent\textbf{Visualization of Perceptual Uncertainty and Similarity Matrix. }  
As shown in Fig.~\ref{fig:e1}, red-circled curb perception results (green polylines) mispredict a left turn as straight, aligning with the highest perceptual uncertainty. The ground truth and predicted results of the PU-aware Similarity Matrix are shown on the right, with the horizontal and vertical axes representing the visual BEV space and map BEV space, respectively. In regions with poor association predictions (upper part), the Gaussian kernel is weaker and less accurate, indicating inferior association predictions for distant areas (due to occlusion).

\noindent\textbf{Consistency between Localization Uncertainty and Localization Quality. }
Fig.~\ref{fig:e2} presents the single-frame inference results for a clip of the nuScenes validation set. It can be observed that the 3DoF errors remain close to zero in most cases, with a positive correlation between uncertainty and localization error. 
Notably, within the blue-boxed region, the longitudinal localization error is generally larger and more unstable, while the longitudinal uncertainty is significantly higher than usual. This phenomenon is caused by the reduced longitudinal constraints in this region, leading to a degeneracy issue. The above results highlight the quality and value of uncertainty prediction.

%% file: tables/o2-nus.tex
\begin{tabular}{lcccccccc}
\toprule
\multirow{2}{*}[-.3em]{Methods}  
& \multicolumn{3}{c}{Position Recall@Xm $\uparrow$}
& \multicolumn{3}{c}{Orientation Recall@X\si{\degree} $\uparrow$}\\ 
\cmidrule(lr){2-4}
\cmidrule(lr){5-7}

&  1m & 3m & 5m & 1\si{\degree} & 3\si{\degree} & 5\si{\degree}\\ 

\midrule

OrienterNet-S   &  5.83 &18.92 &52.83& 32.13 & 41.56 & 65.63 \\
MapLocNet-S  & 8.96 & - & 64.57 & 40.36 & - & 89.66 \\
Ours-S  & \textbf{23.98} & \textbf{39.43} & \textbf{68.73} & \textbf{44.40} & \textbf{70.02} & \textbf{91.39}  \\

\midrule

U-BEV\cite{camiletto2024u}-M    & 16.89 & 41.60 & 71.33 & - & - & -  \\
MapLocNet-M  & 20.10 & - & 77.70 & 58.61 & - & 96.23 \\

Ours-M  & \textbf{34.53} & \textbf{56.21} & \textbf{82.21} & \textbf{60.32} & \textbf{86.28} & \textbf{97.00}  \\

\bottomrule
\end{tabular}

%% file: tables/a1.tex
\begin{tabular}{l|cc|cccc}
\toprule
\multirow{2}{*}[-.2em]{ } & \multirow{2}{*}[-.0em]{\makecell{Assoc. }}  & \multirow{2}{*}[-.0em]{\makecell{Reg. }} & \multirow{2}{*}[+.5em]{\makecell{Lat. $\downarrow$}} & \multirow{2}{*}[+.5em]{\makecell{Lon. $\downarrow$}} & \multirow{2}{*}[+.5em]{\makecell{Ori. $\downarrow$}} \\

& & &  MAE(m) &  MAE(m)&  MAE(\si{\degree}) \\
\midrule
S1   &\checkmark & \checkmark  &  0.040 & 0.140  & 0.075\\
S2   &  &   \checkmark     &  0.058 & 0.157  & 0.108\\
S3   & \checkmark     &      &  0.047 & 0.149  & 0.087\\
S4   &  &                    &  0.064 & 0.160  & 0.360\\

\bottomrule
\end{tabular}

%% file: tables/a2.tex
\begin{tabular}{l|ccc|cccc}
\toprule
\multirow{2}{*}[-.0em]{ } & \multirow{2}{*}[-.0em]{\makecell{LA. }} & \multirow{2}{*}[-.0em]{\makecell{GA. }}  & \multirow{2}{*}[-.0em]{\makecell{PU. }} & \multirow{2}{*}[+.5em]{\makecell{Lat. $\downarrow$}} & \multirow{2}{*}[+.5em]{\makecell{Lon. $\downarrow$}} & \multirow{2}{*}[+.5em]{\makecell{Ori. $\downarrow$}} \\

& & & & MAE(m) &  MAE(m)&  MAE(\si{\degree}) \\
\midrule
S1   &\checkmark &\checkmark & \checkmark  &  0.040 & 0.140  & 0.075\\
S2   & \checkmark &  \checkmark  &         &  0.050 & 0.149  & 0.094\\
S3   & &  \checkmark  &   \checkmark       &  0.043 & 0.144  & 0.081\\

\bottomrule
\end{tabular}

%% file: tables/a3.tex
\begin{tabular}{l|cccccc}
\toprule
& \multicolumn{2}{c}{Lat. $\downarrow$}
& \multicolumn{2}{c}{Lon. $\downarrow$}
& \multicolumn{2}{c}{Ori. $\downarrow$}\\ 
\cmidrule(lr){2-3}
\cmidrule(lr){4-5}
\cmidrule(lr){6-7}
& M.(m) & R.(m) &  M.(m) & R.(m) & M.(\si{\degree}) & R.(\si{\degree})\\ 
\midrule
baseline   & 0.040 & 0.049 & 0.140 & 0.158 & 0.075 & 0.089 \\
w/o PDP.   & 0.042 & 0.054 & 0.144 & 0.184 & 0.079 & 0.099\\ 

\bottomrule
\end{tabular}

%% file: sec/5_conclusion.tex
\section{Conclusion}
This paper presents a novel uncertainty-aware localization network that integrates the advantages of association and registration, demonstrating strong performance across various localization tasks. We thoughtfully design global and local association constraints, where the association is guided by perception uncertainty. Based on the pose distribution and uncertainty estimated by the above module, we perform pose registration to obtain precise localization. Our approach shows significant advantages over existing methods across multiple datasets. Future work will focus on optimizing localization accuracy in extremely challenging scenarios (e.g., with very limited observations) while enhancing the model's generalization ability to achieve a unified model capable of supporting diverse datasets and various maps.